\documentclass[fleqn,10pt]{wlscirep}
\usepackage[utf8]{inputenc}
\usepackage[T1]{fontenc}
\usepackage{verbatim}

\title{Weighted Knowledge Distillation for Semi-Supervised Segmentation of Maxillary Sinus in Panoramic X-ray Images}

\author[1]{Juha Park}
\author[1]{Jiho Choi}
\author[2,3,4]{Jong Pil Yun} 
\author[5]{Yong Chan Park}
\author[5]{Han-Gyeol Yeom}
\author[5]{Byung Do Lee}
\author[1,*]{Sang Jun Lee}

\affil[1]{Division of Electronics and Information Engineering, College of Engineering, Jeonbuk National University, 567, Baekje-daero, Deokjin-gu, 54896, Jeonju, Republic of Korea}
\affil[2]{Manufacturing AI Research Center, Korea Institute of Industrial Technology (KITECH), Incheon 21999, Republic of Korea}
\affil[3]{KITECH School, University of Science and Technology, Cheonan 31056, Republic of Korea}
\affil[4]{The Graduate School of Advanced Imaging Science, Multimedia \& Film, Chung-Ang University, Seoul, Republic of Korea}
\affil[5]{Department of Oral and Maxillofacial Radiology and Wonkwang Dental Research Institute, College of Dentistry, Wonkwang University, 460, Iksan-daero, Iksan-si, 54538, Republic of Korea}
\affil[*]{sj.lee@jbnu.ac.kr}

\begin{abstract}
Accurate segmentation of maxillary sinus in panoramic X-ray images is essential for dental diagnosis and surgical planning; however, this task remains relatively underexplored in dental imaging research.
Structural overlap, ambiguous anatomical boundaries inherent to two-dimensional panoramic projections, and the limited availability of large scale clinical datasets with reliable pixel-level annotations make the development and evaluation of segmentation models challenging.
To address these challenges, we propose a semi-supervised segmentation framework that effectively leverages both labeled and unlabeled panoramic radiographs, where knowledge distillation is utilized to train a student model with reliable structural information distilled from a teacher model. 
Specifically, we introduce a weighted knowledge distillation loss to suppress unreliable distillation signals caused by structural discrepancies between teacher and student predictions.
To further enhance the quality of pseudo labels generated by the teacher network, we introduce SinusCycle-GAN which is a refinement network based on unpaired image-to-image translation. 
This refinement process improves the precision of boundaries and reduces noise propagation when learning from unlabeled data during semi-supervised training.
To evaluate the proposed method, we collected clinical panoramic X-ray images from 2,511 patients, and experimental results demonstrate that the proposed method outperforms state-of-the-art segmentation models, achieving the Dice score of 96.35\% while reducing boundary error.
The results indicate that the proposed semi-supervised framework provides robust and anatomically consistent segmentation performance under limited labeled data conditions, highlighting its potential for broader dental image analysis applications.
\end{abstract}

\begin{document}

\flushbottom
\maketitle
\thispagestyle{empty}

\section*{Introduction}

The maxillary sinus is the largest paranasal sinus within the maxilla and has a hollow, pyramidal shape.
It contributes to the regulation of the humidity and temperature of the inspired air, generates resonance during speech, and facilitates mucociliary clearance.
In particular, its size and morphology vary substantially between individuals, and its close proximity to posterior maxillary teeth requires detailed anatomical information in this region, which is particularly important for dental interventions.
Accurate identification of the location of the maxillary sinus is essential for various surgical procedures involving teeth adjacent to the maxillary sinus floor, including implant placement and tooth extraction.
Specifically, implant placement in the posterior maxilla is often challenging due to limited vertical bone height and poor bone quality, and there is a risk of perforating the sinus floor when the maxillary sinus is in proximity~\cite{Sala:2024}.
Therefore, an accurate assessment of the position and morphology of the maxillary sinus is required to achieve successful surgical outcomes.

The preoperative assessment of the maxillary sinus can be performed using panoramic radiography, periapical radiography, and cone-beam computed tomography (CBCT).
Among these modalities, panoramic radiography has remained the most widely used imaging tool in clinical practice due to its cost-effectiveness, accessibility, and low radiation exposure.
It provides clinically valuable information for planning various surgical procedures, such as maxillary sinus elevation.
While CBCT offers superior three-dimensional visualization through multidirectional image reconstruction, its routine clinical application is constrained by high cost, increased radiation exposure, and longer scanning time.
Moreover, panoramic radiography continues to provide sufficient anatomical information for most dental procedures, despite some limitations in delineating the maxillary sinus anatomy due to superposition of anatomical structures.
Therefore, it is essential to evaluate the clinical effectiveness of panoramic radiography for assessing maxillary sinus anatomy in clinical practice~\cite{Rushton:1996, malina2015evaluation}.

Radiographic interpretation of anatomical structures within the maxillary sinus on panoramic radiographs is important for the initial diagnosis in dental implant planning.
Panoramic radiography offers a comprehensive overview of the maxillary region, allowing the simultaneous evaluation of the teeth, jaws, and maxillary sinus in a single scan.
Such simultaneous visualization allows clinicians to acquire basic anatomical information about the internal structure of the maxillary sinus, including the identification of sinus septa~\cite{malina2015evaluation}.
Furthermore, panoramic radiography is associated with lower costs and reduced radiation exposure compared to CBCT.
The short acquisition time and simple examination procedure improve patient comfort and acceptability.
The combination of low cost, reduced radiation exposure, and efficient image acquisition establishes panoramic radiography as a valuable first-line imaging modality to evaluate maxillary sinus anatomy in treatment planning.

In dentistry and other clinical fields, segmentation of anatomical structures is predominantly performed using manual or semi-automatic approaches by trained experts.
Manual segmentation is time-consuming and labor-intensive for accurate analysis, and the results are highly dependent on domain knowledge and experience.
Consequently, the results may vary between different experts as well as for the same expert at different time points, even when analyzing identical images.
Although semi-automatic segmentation techniques can improve efficiency, they still require manual adjustments that may introduce errors~\cite{Morgan:2022}.
These limitations in reliability and consistency make manual and semi-automatic approaches less suitable for clinical applications requiring standardized analysis.
In recent years, automatic segmentation has emerged as a promising method, with researchers demonstrating expert-level accuracy across various medical imaging applications.

Recently, segmentation has been studied in medical imaging and is rapidly being adopted in dental imaging analysis.
The encoder–decoder architecture including U-Net~\cite{Ronneberger:2015} and its variants has achieved high accuracy and computational efficiency in segmentation.
Segmentation models have been utilized in diverse clinical applications such as periapical lesion detection, tooth boundary recognition, root length estimation, and periodontal tissue analysis with reported expert-level performance~\cite{Koch:2019, BaHattab:2023, Hiraiwa:2019, Krois:2019}.
In panoramic radiography, several models have been introduced for tasks such as individual tooth segmentation, detection of periapical lesions, and extraction of jawbone contours, demonstrating their potential utility as clinical assistance tools~\cite{BonfantiGris:2025, Poedjiastoeti:2018}.
These approaches outperformed traditional image processing techniques in accuracy and consistency and exhibited enhanced robustness to diverse data and noise.
Nevertheless, the majority existing studies focus on teeth and adjacent soft tissues, whereas segmentation research on complex anatomical structures of the maxillary sinus remains scarce.
The maxillary sinus presents additional challenges due to its mixed air–bone composition and indistinct boundaries.
Accordingly, there is a need for segmentation approaches capable of capturing the complex anatomical features of the maxillary sinus in imaging data.
In this context, this study aims to develop and validate a novel deep learning algorithm for efficient identification and automatic segmentation of the maxillary sinus in panoramic radiography.

Automatic maxillary sinus segmentation remains underexplored and most existing methods rely on clearly defined three-dimensional anatomical information, such as computed tomography (CT) images.
In contrast, accurate delineation of the maxillary sinus from panoramic radiographs requires learning meaningful anatomical structures from incomplete two-dimensional information affected by structural overlap and distortion.
Moreover, pixel-level annotation demands substantial time and expert effort, making the construction of large-scale labeled datasets particularly challenging.
To overcome these limitations, we propose a semi-supervised segmentation framework for automatic identification and segmentation of the maxillary sinus in panoramic radiographs by effectively leveraging a small amount of labeled data together with a large volume of unlabeled data.
We adopt a teacher-student framework based on knowledge distillation (KD) as the core learning structure and define a weighted knowledge distillation loss to ensure that only reliable knowledge is transferred from the teacher model.
Although recent foundation models pretrained on large scale medical datasets provide strong representational capacity, their predictions may still exhibit boundary inaccuracies and noise when applied to panoramic radiographs, owing to domain specific characteristics.
To address this issue, we further introduce the SinusCycle-GAN network, which refines pseudo labels from the foundation model to enhance structural reliability and improve segmentation performance on unlabeled data.
To demonstrate the effectiveness of the proposed method, we collected clinically acquired panoramic X-ray images from 2,511 patients and experimental results indicate that the proposed method outperforms existing segmentation models.
The main contributions of this paper are summarized as follows:
\begin{itemize}
\item We propose a semi-supervised framework for maxillary sinus segmentation in panoramic X-ray images that effectively leverages large amounts of unlabeled data.
\item We introduce weighted knowledge distillation loss that adaptively regulates the distillation process according to structural consistency between teacher and student predictions.
\item We develop SinusCycle-GAN, a refinement network designed to enhance the quality of pseudo labels generated by the teacher model in the proposed semi-supervised framework.
\item We conduct extensive experiments on clinical panoramic X-ray images and demonstrate state-of-the-art segmentation performance.
\end{itemize}

\section*{Related work}

\subsection*{Analysis of panoramic radiography image}
Early dental image analysis was primarily based on traditional image processing methods.
Huang et al.~\cite{Huang:2021} applied multiple image-enhancement methods, such as sharpening, histogram equalization, and flattening correction, to segment teeth from panoramic radiographs.
This approach improved segmentation performance through image enhancement; however, segmentation artifacts were observed due to difficulties in robustly identifying inter-tooth spacing features.
Majanga et al.~\cite{Majanga:2021} extracted tooth regions of interest by applying morphological operations and eight-directional connected-component analysis. In this pipeline, image processing methods were applied during preprocessing, which directly affected the final performance of the deep learning system.
These conventional methods required expert-dependent manual adjustment, and they were criticized for limited consistency and difficulty in ensuring reproducibility.

To address these limitations, many deep learning approaches for panoramic radiographs have recently been proposed in dentistry~\cite{Cha:2021}.
Reported applications include semantic segmentation, object detection, and instance segmentation.
Koch et al.~\cite{Koch:2019} performed segmentation of dental panoramic radiographs using a U-Net-based fully convolutional network.
Kwon et al.~\cite{Kwon:2020} proposed a novel deep learning framework to automatically diagnose odontogenic cysts and tumors from panoramic radiographs in both jaws.
Their method integrates data augmentation to support detection and classification across multiple disease types.
Lee et al.~\cite{Lee:2020} introduced a fully deep learning convolutional neural network algorithm for automatic tooth segmentation in panoramic radiographs.
Despite the substantial success of deep learning models in medical imaging, its application to dental image remains relatively underexplored~\cite{Murata:2019}.

\subsection*{Knowledge distillation for segmentation}
KD is a machine learning framework that transfers knowledge from a large, pretrained teacher model to a smaller and more efficient student model, thereby improving the student’s performance.
The KD framework proposed by Hinton et al.~\cite{Hinton:2015} optimizes the student model using soft target probability distributions produced by the teacher model, rather than conventional hard labels.
Zheng et al.~\cite{Zheng:2019} proposed a KD method that distills knowledge from multiple regions of an image to enhance fine-grained visual classification performance.
By integrating fine-grained local features into a single-stream architecture, their method achieved superior performance compared with part-based ensemble models.
These early KD methods primarily focused on classification tasks; however, the ability of KD to transfer structural and relational knowledge has motivated its extension to dense prediction problems such as semantic segmentation.

In the field of segmentation, a variety of KD methods have been proposed to improve the performance of the model.
CIRKD~\cite{Yang:2022} is a KD method that transfers global relational knowledge, specifically pixel-to-pixel and pixel-to-region relationships, from the teacher model to the student model, in contrast to conventional KD methods that primarily transfer pixel-level structural information within an image.
BPKD~\cite{Liu:2024BPKD} is a KD framework that explicitly distinguishes boundary regions and interior regions, which are the most challenging areas in segmentation tasks, and applies different loss functions and knowledge-transfer strategies to each region.
In addition, Liang et al.~\cite{Liang:2024RDD} incorporated the concept of relative sample difficulty learned by the teacher and student models, proposing a KD method that provides stronger feedback for samples that are more difficult to learn.

Recently, KD frameworks have also been applied to the medical imaging domain, and consistent segmentation performance has been reported even under limited and low-quality annotated data conditions.
Dong et al.~\cite{Dong:2025SIKD} proposed a KD method in which the teacher model learns shape–intensity prior information and transfers this knowledge to the student model, thereby improving the robustness and generalization performance of medical image segmentation models.
This method effectively integrates prior information on morphological structure and image intensity, resulting in significant segmentation performance gains across five medical image segmentation tasks involving different imaging modalities.
Furthermore, Hu et al.~\cite{Hu:2020MMKD} proposed a KD method that transfers knowledge from a teacher model using multi-modal inputs to a mono-modal student model, and demonstrated its effectiveness in medical imaging applications, particularly for brain tumor segmentation.
KD is regarded as a promising framework for improving performance in medical image segmentation by effectively transferring knowledge from large scale models.
Motivated by these studies, this work proposes a semi-supervised framework with knowledge distillation that improves segmentation performance even with a limited amount of annotated data.

\begin{figure}[ht]
\centering
\includegraphics[width=\linewidth]{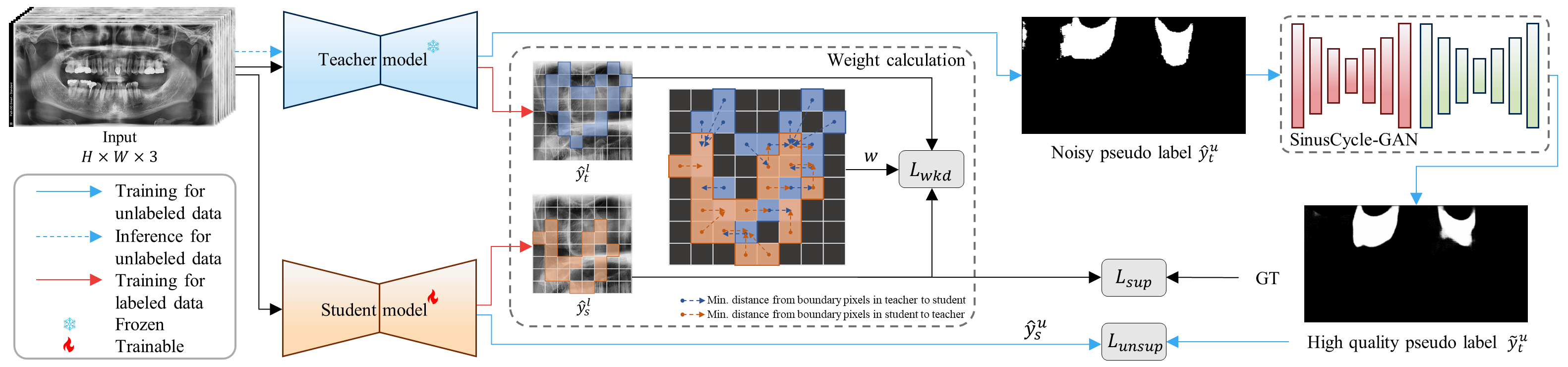}
\caption{Overview of the proposed method. Panoramic X-ray images are fed into both the teacher and student models. The teacher model is trained using only labeled data, whereas the student model is trained using both labeled and unlabeled data. After the teacher model is trained, it generates pseudo labels for the unlabeled data. These pseudo labels are refined by SinusCycle-GAN and subsequently used to train the student model through a weighted knowledge distillation loss.}
\label{fig:main}
\end{figure}

\begin{figure}[ht]
\centering
\includegraphics[width=\linewidth]{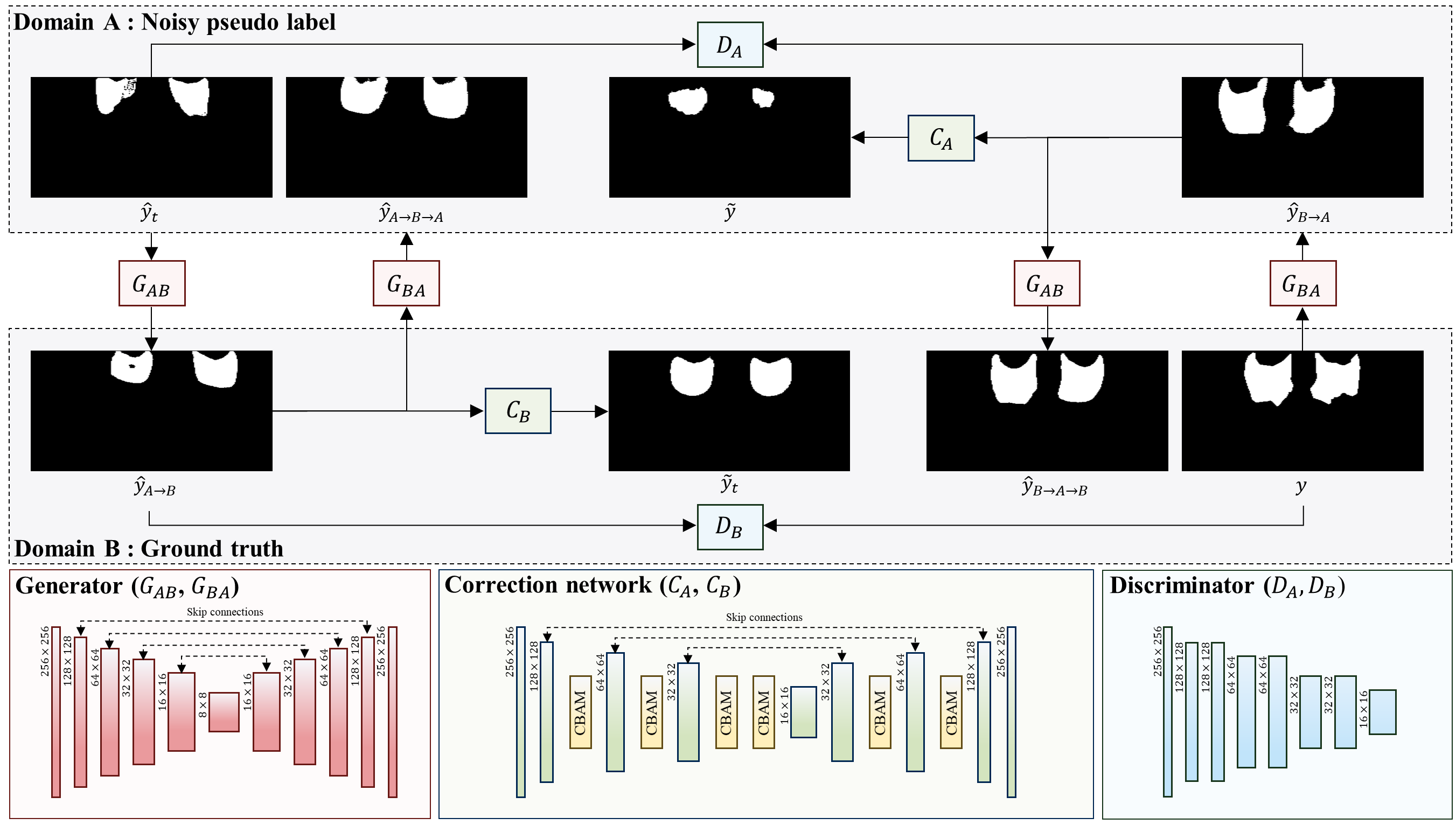}
\caption{Architecture of the proposed SinusCycle-GAN for pseudo label refinement. The network is based on an unpaired image-to-image translation framework and incorporates correction network with an encoder-decoder architecture with CBAM to enhance boundary accuracy and suppress noise in initial pseudo labels $\hat{y}_t$. The refined output is $\tilde{y}_t$ and corresponds to the high-quality pseudo label used to train the student model.}
\label{fig:sub}
\end{figure}    

\section*{Proposed method}

In this paper, we propose a semi-supervised segmentation framework for accurate maxillary sinus segmentation in panoramic X-ray images.
The overall architecture of the proposed approach is illustrated in Figure~\ref{fig:main}.
Our method adopts a teacher–student learning architecture in which both teacher and student segmentation networks are based on MedSAM, a specialized foundation model for medical image segmentation.
To ensure reliable knowledge transfer, we introduce a weighted knowledge distillation strategy that quantifies structural discrepancies between teacher and student predictions using the 95th percentile Hausdorff distance (HD95) and adaptively suppresses unreliable distillation signals.
We employ SinusCycle-GAN as a refinement network to address performance degradation caused by inaccurate and noisy pseudo labels by correcting boundary distortions and suppressing spurious noise in teacher-generated predictions.
Given a 2D panoramic X-ray image provided as input at a resolution of $1024 \times 1024$, the teacher model is trained on labeled data, while the student model is trained using both labeled and unlabeled data through weighted knowledge distillation and refined pseudo labels.
The final segmentation output is produced at a resolution of $2048 \times 1024$.
By jointly leveraging a small amount of labeled data and a large volume of unlabeled data, the proposed approach achieves improved segmentation performance with enhanced boundary accuracy.
Detailed explanations for each component are provided in the subsequent sections.

\subsection*{SinusCycle-GAN}
We introduce SinusCycle-GAN as a refinement network to address the degradation in student model performance caused by noisy pseudo labels.
During training, both teacher and student models are based on MedSAM~\cite{medsam}.
The performance of the student model on unlabeled data critically depends on the quality of pseudo labels generated by the teacher model; therefore, ensuring high-quality pseudo labels is essential.
In practice, the initial pseudo labels often contain various errors, including boundary distortion, omission of small structures, and discontinuous noise, which can lead to performance degradation.
To mitigate this issue, we introduce SinusCycle-GAN to improve the quality of pseudo labels.
SinusCycle-GAN adopts the structure of CycleGAN~\cite{cyclegan}.
To further enhance the quality of the generated masks, a correction network is incorporated after the generation process to enable more accurate and robust refinement.
As a result, SinusCycle-GAN generates a high-quality pseudo label as its final output.

Figure~\ref{fig:sub} describes the architecture of the proposed SinusCycle-GAN.
We define domain~$A$ as the set of initial noisy pseudo labels, denoted as $\hat{y}_t \in A$, and domain~$B$ as the set of ground truth masks from labeled data, denoted as $y \in B$.
To train SinusCycle-GAN, we constructed two generators, two discriminators, and two correction networks to perform unpaired image-to-image translation.
The first generator $G_{AB}$ takes $\hat{y}_t$ as input and generates a translated mask $\hat{y}_{A \to B}$, that resembles the ground truth mask $y$.
The first discriminator $D_{B}$ is trained to distinguish between $\hat{y}_{A \to B}$ and $y$.
As the adversarial training proceeds, the generator $G_{AB}$ produces masks that are more similar to $y$.
Following the CycleGAN framework, we defined the adversarial loss for unpaired image-to-image translation.
For stable optimization, we adopt a least-squares GAN~\cite{LSGAN} formulation, and the adversarial loss used for training $G_{AB}$ and $D_{B}$ is defined as
\begin{equation}
L_{adv}(G_{AB}, D_{B}, A, B) = \mathbb{E}_{y\sim p_{data}(y)}\!\left[(D_{B}(y)-1)^2\right] + \mathbb{E}_{\hat{y}_t\sim p_{data}(\hat{y}_t)}\!\left[D_{B}(G_{AB}(\hat{y}_t))^2\right],
\end{equation}
where $p_{data}(\hat{y}_t)$ and $p_{data}(y)$ are the distributions of the domain~$A$ and domain~$B$.
Similarly, the other generator $G_{BA}$ maps samples from domain~$B$ to domain~$A$.
The second discriminator $D_{A}$ is trained to distinguish between $\hat{y}_t$ and generated masks $\hat{y}_{B \to A}$.
Another adversarial loss for training $G_{BA}$ and $D_{A}$ is defined as 
\begin{equation}
L_{adv}(G_{BA}, D_{A}, B, A) = \mathbb{E}_{\hat{y}_t \sim p_{data}(\hat{y}_t)}\! \left[(D_{A}(\hat{y}_t)-1)^2 \right] + \mathbb{E}_{y \sim p_{data}(y)}\! \left[D_{A}(G_{BA}(y))^2 \right].
\end{equation}
As adversarial loss alone is insufficient to preserve semantic correspondence between the input and output, a cycle consistency loss is introduced to ensure that the translated images can be reconstructed to the original domain, thereby encouraging the preservation of structural information.
The cycle consistency loss is defined as
\begin{equation}
L_{cycle}(G_{AB}, G_{BA}) = \mathbb{E}_{\hat{y}_t \sim p_{data}(\hat{y}_t)} \left[\left\| G_{BA}(G_{AB}(\hat{y}_t)) - \hat{y}_t \right\|_{1} \right] + \mathbb{E}_{y \sim p_{data}(y)} \left[ \left\| G_{AB}(G_{BA}(y)) - y \right\|_{1} \right].
\end{equation}

Cycle consistency provides global structural guidance, but additional refinement is required to capture fine-grained boundary details.
To address this, an additional correction network is incorporated into the overall framework.
The architecture of the correction network is shown in the lower center of Figure~\ref{fig:sub}.
The correction network is based on an encoder–decoder structure.
The encoder transforms the input image into an abstract feature representation, and the decoder generates an output image that is similar to the original image while capturing refined and discriminative features.

The encoder first performs 4 stages of downsampling to extract hierarchical features from the input image.
To enable stable training in deep network structures, residual connections are introduced to alleviate vanishing gradients and to mitigate information loss.
During this process, convolutional block attention module (CBAM)~\cite{cbam} is selectively applied after intermediate downsampling stages, excluding the input layer and the bottleneck layer.
This design emphasizes informative features along both channel and spatial dimensions for feature maps with varying resolutions and channel sizes, allowing the network to focus on structurally meaningful regions.
The decoder restores the original spatial resolution of the mask through 4 stages of upsampling based on the encoded feature maps.
To effectively preserve fine structural details that may be lost during downsampling, features from the encoder are integrated via skip connections.
CBAM is also selectively applied after intermediate upsampling layers in the decoder, where they enhance important channel-wise and spatial information from the combined representations of upsampled features and skip-connected features.
As a result, boundary regions and structurally critical features are reconstructed more sharply.
For training the correction network, we employ the binary cross-entropy (BCE) loss defined in Equation.~\ref{eq:bce}, which is suitable for pixel-level binary classification tasks.
\begin{equation}
L_{{BCE}}(a, b)
= - \left[ b \log(\sigma(a)) + (1-b)\log(1-\sigma(a)) \right],
\label{eq:bce}
\end{equation}
where $a$ and $b$ denote the pixel-wise logit predictions and binary labels, and $\sigma(\cdot)$ is the sigmoid function.
Based on the BCE loss defined above, the correction loss used to train the correction network is defined as
\begin{equation}
L_{corr}(G_{AB}, G_{BA}, C_{B}, C_{A}) = \mathbb{E}_{\hat{y}_t\sim p_{data}(\hat{y}_t)} \!\left[L_{BCE}(C_{B}(G_{AB}(\hat{y}_t)), y) \right] + \mathbb{E}_{y\sim p_{data}(y)} \!\left[L_{BCE}(C_{A}(G_{BA}(y)), \hat{y}_t) \right],
\end{equation}
where $C_B$ and $C_A$ denote the correction networks used to refine the masks generated by mapping from domain~$A$ to domain~$B$ and from domain~$B$ to domain~$A$, respectively.

The total loss function used to train SinusCycle-GAN is fomulated as the combination of the adversarial losses, the cycle consistency loss, and the correction loss, and it is defined as
\begin{equation}
L_{total}^{sinus} = L_{adv}(G_{AB}, D_{B}, A, B) + L_{adv}(G_{BA}, D_{A}, B, A) + \lambda L_{cycle}(G_{AB}, G_{BA}) + L_{corr}(G_{AB}, G_{BA}, C_B, C_A),
\end{equation}
where $\lambda$ is a hyper-parameter that controls the contribution of the cycle consistency loss, and we utilized 10 in experiments.
The final output of SinusCycle-GAN is denoted as the high-quality pseudo label $\tilde{y}_{t}^{u}$.

\subsection*{Semi-supervised segmentation based on knowledge distillation}
In the medical imaging domain, the acquisition of large scale labeled datasets remains challenging, as data availability is constrained by patient privacy considerations and annotation requires substantial time and effort from experienced medical experts.
To effectively leverage both labeled and unlabeled data, this study employs a teacher–student knowledge distillation framework with semi-supervised segmentation.
Within this framework, reliable knowledge is distilled from a teacher model into a student model using both labeled and unlabeled data.
The teacher model is first trained using labeled data with ground truth masks.
The trained teacher is then frozen and used to generate pseudo labels for unlabeled data.
Through this process, the student model learns from both labeled and unlabeled data, substantially improving data utilization efficiency throughout training.

In medical image segmentation, predictions with significant structural discrepancies are unsuitable for knowledge distillation; therefore, it is essential to assess the discrepancy between teacher and student predictions.
We introduce a weighting strategy based on the Hausdorff distance between the predictions of the two models to estimate the image-level reliability.
The Hausdorff distance measures boundary errors between two sets by computing the maximum distance from a point in one set to the closest point in the other set, but it is highly sensitive to outlier pixels.
To mitigate this sensitivity, we employ HD95.
The measured value is used as a weighting coefficient in weighted knowledge distillation loss $L_{wkd}$.
This strategy prevents incorrect distillation signals that arise from structurally inconsistent predictions from degrading the training process.

In this study, we define a semi-supervised training objective that combines supervised learning on labeled data with weighted knowledge distillation between the teacher and student models, and further incorporates unsupervised learning based on pseudo labels for unlabeled data.
For labeled data, the student model is optimized using a supervised segmentation loss that directly enforces agreement with ground truth annotations.
This loss is defined as the sum of the Dice loss and the BCE loss, jointly accounting for image-level overlap accuracy and pixel-level classification accuracy.
Dice loss optimizes region overlap between the predicted mask and the ground truth mask and is robust to class imbalance, which is common in medical imaging where target structures occupy only a small region of the image.
BCE loss helps regularize the pixel-level probability estimation, but can become biased toward the background class under severe class imbalance.
By combining these two loss functions, the proposed approach improves both global structural alignment and local pixel-level accuracy.
Let $z_s^{l}$ denote the output logit map of the student model for labeled data, and let $p_s^{l} = \sigma(z_s^{l})$ denote the corresponding probability map.
\begin{equation}
L_{Dice} = 1 - 
\frac{ 2 \sum \left( p_s^{l} \cdot y \right) }
{ \sum \left( p_s^{l} \right)^2 + \sum \left( y \right)^2 } .
\end{equation}
\begin{equation}
L_{sup}
= L_{Dice}
+ L_{{BCE}}(z_{s}^{l}, y),
\end{equation}
where $y$ denotes the corresponding binary ground truth mask and we employ the BCE loss defined in Equation.~\ref{eq:bce}.
The Dice and BCE loss are computed on a pixel-level basis for each sample and averaged over the batch dimension.

To ensure reliable knowledge transfer, we introduce a weighted knowledge distillation loss $L_{wkd}$ that incorporates an image-level weighting coefficient to suppress structurally inconsistent teacher supervision.
During the distillation process, a temperature parameter $T > 1$ is incorporated into the softmax function to compute soft predictions.
Let $z_t^{l}$ denote the output logit map of the teacher model for labeled data.
The softened probability distributions $q_s$ and $q_t$ of the student and teacher models are defined as
\begin{equation}
q_s = \mathrm{softmax} \left( \frac{\mathbf{z}_s^{l}}{T} \right), \quad
q_t = \mathrm{softmax} \left( \frac{\mathbf{z}_t^{l}}{T} \right).
\end{equation}
Here, $\mathbf{z}_s^{l}$ and $\mathbf{z}_t^{l}$ denote the student and teacher logit vectors for binary classification. 
Specifically, $\mathbf{z}_s^{l} = (0, z_s^{l})$ and $\mathbf{z}_t^{l} = (0, z_t^{l})$, where the first component corresponds to the background class.
The conventional knowledge distillation loss is defined as the Kullback--Leibler (KL) divergence between the soft predictions of the teacher and student models.
However, in medical image segmentation, when the predictions of the teacher and student models differ significantly in terms of structural consistency, incorrect distillation knowledge may degrade the learning process.
To alleviate this issue, we quantify the structural discrepancy between the teacher and student predictions using HD95 and introduce a weighting coefficient into the conventional knowledge distillation loss.
Let $U$ and $V$ denote the sets of pixel coordinates extracted from the prediction masks of the teacher model and student model.
$U = \{u_1, \ldots, u_N\} \subset \hat{y}_t^{l}, \quad V = \{v_1, \ldots, v_M\} \subset \hat{y}_s^{l}$, where  $\hat{y}_t^{l}$ and $\hat{y}_s^{l}$ denote the prediction masks of the teacher and student model for labeled data.
Then, the Hausdorff distance is defined as
\begin{equation}
\mathrm{HD}(U, V) = \max\!\left( h(U,V), h(V,U) \right),
\quad \text{ where }\; h(U,V) = \max_{u \in U} \min_{v \in V} \lVert u - v \rVert.
\end{equation}
To reduce sensitivity to outliers, we use the HD95, which is computed by replacing the maximum operator in $h(\cdot,\cdot)$ with the 95th percentile of the corresponding pixel distance distribution.
Accordingly, $\mathrm{HD95}_i$ denotes the $\mathrm{HD95}$ value computed between the teacher and student prediction masks for the $i$-th sample.
The weighting coefficient for the \(i\)-th sample, denoted as \(w_{i}\), is defined as a function of the structural discrepancy between the teacher and student predictions.
This coefficient modulates the contribution of each sample to the weighted knowledge distillation loss.
\begin{equation}
w_{i} = \frac{\exp\!\left(-\mathrm{HD95}_i / \tau\right)} {\max_{j} \exp\!\left( -\mathrm{HD95}_j / \tau\right)}.
\end{equation}
Here, $\tau$ is a hyper-parameter that controls the scale of the HD95 value. 
The index $j \in \{1,\ldots,B\}$ denotes all samples within the mini-batch and the maximum operation is used to normalize the weighting coefficients across the batch.
Consequently, the weighting coefficient $w_i$ is bounded in the range $(0, 1]$.
Samples with low structural discrepancies between the teacher and student predictions are assigned larger weights, leading to greater knowledge distillation.
The proposed weighted knowledge distillation loss is defined as
\begin{equation}
L_{{wkd}} = \frac{T^{2}}{B} \sum_{i=1}^{B} w_{i}\, \mathrm{KL}\!\left( q_{t,i} \,\|\, q_{s,i} \right),
\end{equation}
where $B$ denotes the batch size, and $T$ is set to 2 in our experiments.

For unlabeled data, learning is guided by a high-quality pseudo label $\tilde{y}_{t}^{u}$ refined by SinusCycle-GAN, where unsupervised loss is calculated only over pixels predicted as foreground by the teacher model, thus emphasizing anatomically relevant regions and reducing the influence of background noise.
Let $\hat{y}_s^{u}$ denote the prediction mask of the student model for unlabeled data.
\begin{equation}
L_{{unsup}} = \frac{\sum_{i} \tilde y_{t,i}^{u}\; L_{BCE}\!\left(\hat y_{s,i}^{u},\, \tilde y_{t,i}^{u}\right)} {\sum_{i} \tilde y_{t,i}^{u}} .
\end{equation}
The total training loss is defined as a weighted sum of the supervised, distillation, and unsupervised loss terms.
\begin{equation}
L_{total} = \alpha L_{sup} + \beta L_{wkd} + \left( 1 - \alpha \right) L_{unsup}.
\end{equation}
Here, $\alpha$ is a hyperparameter that controls the relative contribution of labeled and unlabeled data during training, and $\beta$ denotes the weighting factor for the weighted knowledge distillation loss.
In our experiments, $\alpha$ and $\beta$ are set to 0.5 and $1 \times 10^{-6}$, respectively.

\section*{Experimental results}

We conducted experiments with the hardware environment including 192-core CPU, 64 GB DDR4 RAM, and an NVIDIA H100 GPU with 80 GB HBM3 memory.
The proposed algorithm was implemented based on Pytorch.
To train the segmentation network, we utilized the AdamW optimizer with the learning rate of 0.00001.
Training was performed for 10 epochs with a batch size of 8.
The performance of the model was evaluated on the training and validation sets during training, and the final results were reported on the test set.
SinusCycle-GAN was trained using the same training settings for 100 epochs with a batch size of 10.

\begin{figure}[ht]
\centering
\includegraphics[width=\linewidth]{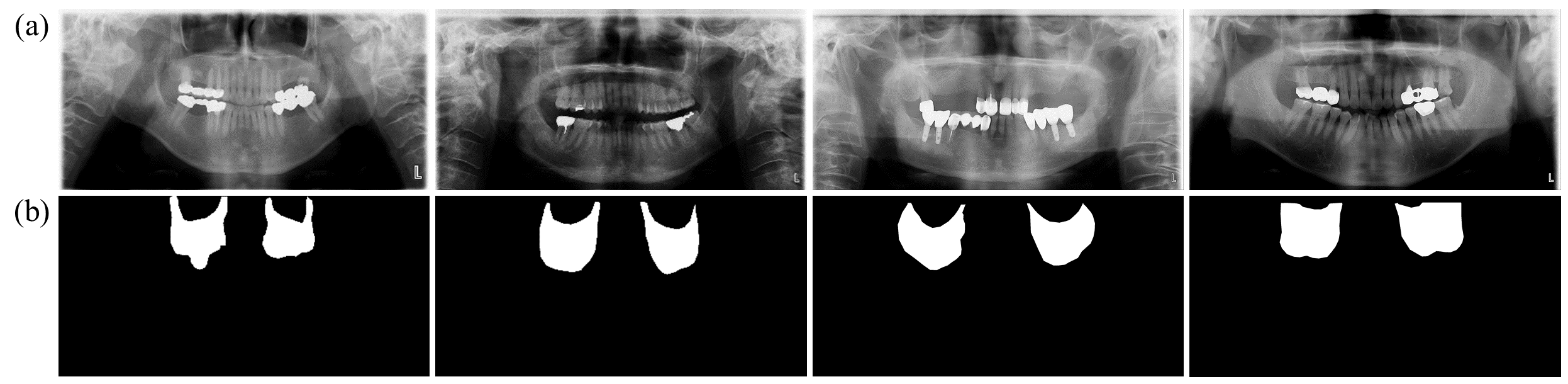}
\caption{Examples of panoramic X-ray images for maxillary sinus segmentation. (a) panoramic X-ray images. (b) ground truth masks annotated by expert.}
\label{fig:data}
\end{figure}

\begin{figure}[ht]
\centering
\includegraphics[width=11cm]{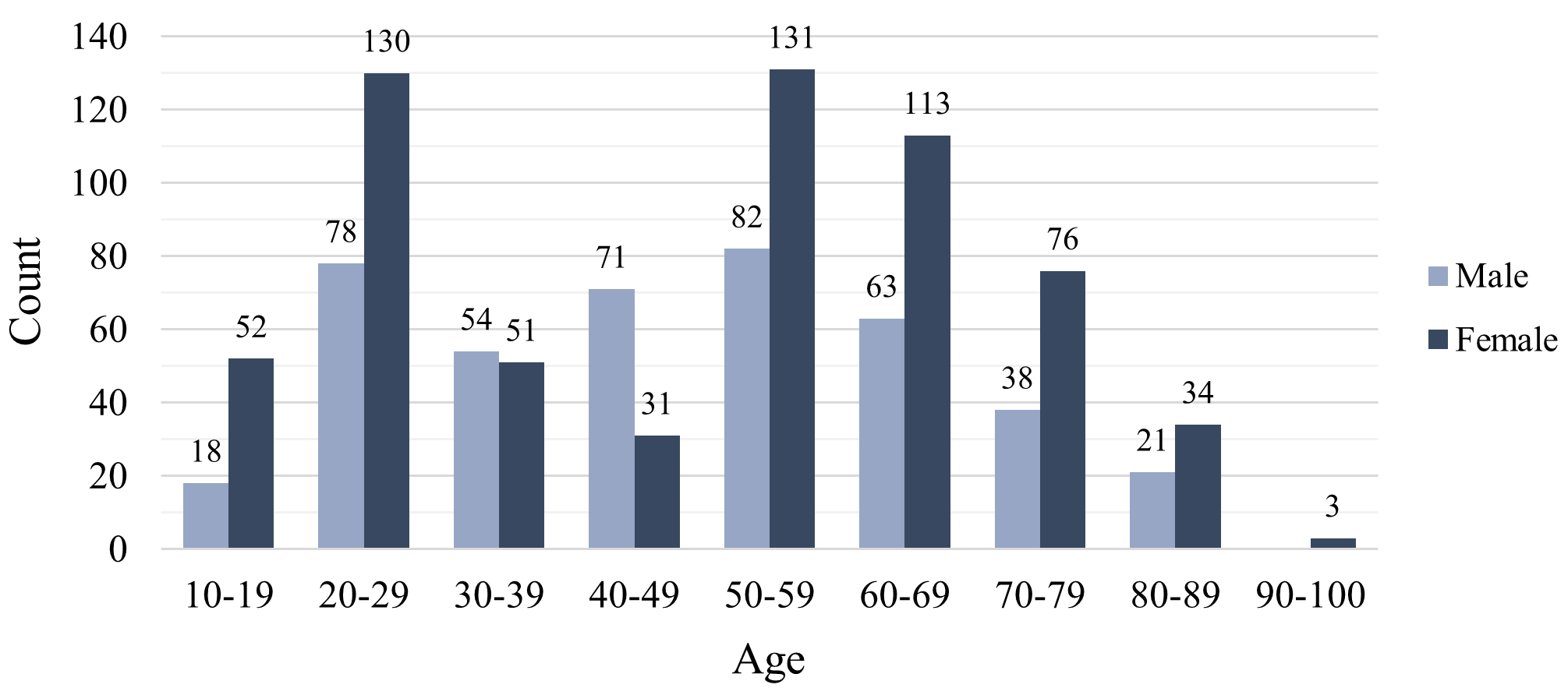}
\caption{Number of subjects by age group and sex in the maxillary sinus clinical dataset.}
\label{fig:chart}
\end{figure}

\subsection*{Dataset Description}
To evaluate the proposed method, we utilized a clinically acquired panoramic X-ray dataset retrospectively collected from patients at a university-affiliated dental hospital and several local dental clinics. 
The study population was randomly sampled from individuals receiving a variety of dental procedures, including impacted third molar extractions, orthodontic assessments, and diagnostic evaluation for dental pain.
Inclusion criteria were restricted to patients with no prior history of maxillofacial surgery and with either normal maxillary sinus mucosa or only mild mucosal thickening observed on panoramic radiographs.
Patients presenting with congenital jaw deformities or documented maxillofacial trauma that could compromise reliable radiographic interpretation were excluded.

As illustrated in Figure~\ref{fig:data}, each sample consists of a panoramic X-ray image and a corresponding label delineating the region of the maxillary sinus.
Manual annotation was performed using the VIA™ annotation platform (VGG Image Annotator, Visual Geometry Group, University of Oxford, UK).
The limits of the maxillary sinus were delineated through an annotation performed by an expert based on a consensus between an oral and maxillofacial radiology specialist and a general dentist.
All annotations were performed using polygon based tools and exported from the VIA interface in CSV format.
Each CSV file contained both the pixel-level annotation information for the maxillary sinus and associated clinical metadata, including patient age, sex, acquisition date, and disease presence.

The complete dataset was divided into training, validation, and test sets comprising 2,091, 210, and 210 images, respectively.
For the semi-supervised learning setting, the training set was further partitioned into labeled and unlabeled subsets.
Among the training samples, 626 images were accompanied by maxillary sinus labels annotated by experts and used as labeled data.
Figure~\ref{fig:chart} illustrates the age and sex distribution of the subjects included in the labeled dataset.
Patient ages ranged from 13 to 92 years.
Among the 1,046 participants, females accounted for 59.4\% and males for 40.6\%.
The remaining 1,465 training images lacked manual annotations and were considered unlabeled data.
Unlabeled images were incorporated into the training process using initial pseudo labels generated by the teacher model within the proposed semi-supervised framework.
The dataset is collected with the approval of the Institutional Review Board of Wonkwang University Dental Hospital (IRB No.: WKDIRB202107-01).
Written informed consent was obtained from all participants or their legal guardians, and all procedures were performed in accordance with relevant guidelines and regulations.

\subsection*{Evaluation Measures}
The proposed method was evaluated using the Dice coefficient, recall, precision, and the HD95 by comparing the predicted regions with the ground truth masks.
Here, true positives (TP) denote pixels correctly predicted as foreground, false positives (FP) denote background pixels incorrectly predicted as foreground, and false negatives (FN) denote foreground pixels incorrectly predicted as background.
The Dice coefficient measures the overlap between the predicted region and the ground truth region, with higher values indicating better segmentation performance.
It is defined as
\begin{equation}
\mathrm{Dice} = \frac{2TP}{2TP + FP + FN}.
\end{equation}
Recall evaluates the proportion of ground truth pixels that are correctly identified by the model, while precision measures the proportion of predicted foreground pixels that are correct.
They are defined as
\begin{equation}
\mathrm{Recall} = \frac{TP}{TP + FN},
\end{equation}
\begin{equation}
\mathrm{Precision} = \frac{TP}{TP + FP}.
\end{equation}
The HD95 measures the boundary distance between the predicted segmentation and the ground truth, where a lower value indicates better segmentation accuracy.

\subsection*{Quantitative Evaluation}
Table~\ref{tab:quantitative_comparison} presents the quantitative comparison of the proposed method with several previous methods.
The proposed method achieved the highest segmentation performance across all evaluation metrics, attaining the Dice score of 96.35\%, recall of 97.34\%, and precision of 95.90\%.
In addition, it yielded the lowest HD95 value of 0.0138, indicating superior boundary alignment with the ground truth masks.
These results demonstrate that the proposed method not only improves region overlap accuracy but also achieves more precise and reliable boundary delineation compared with existing methods.
Notably, the proposed method achieves improved segmentation performance while maintaining identical parameter count and FLOPs to MedSAM, indicating that the performance gains are obtained without any increase in model size or computational complexity.

Compared to MedSAM, the proposed method improved the Dice score by 0.33\%, recall by 0.86\%, and precision by 0.26\%.
Furthermore, the HD95 value was reduced from 0.0146 to 0.0138, corresponding to a 5.5\% decrease and indicating more accurate boundary localization.
Compared to TransUNet, the proposed method demonstrated a more balanced performance with improved precision and a markedly lower HD95.
These consistent improvements across overlap-based, pixel-level, and boundary-based metrics indicate that the proposed framework effectively enhances both segmentation accuracy and structural fidelity in panoramic maxillary sinus segmentation.

\begin{table}[h]
\centering 
\caption{Quantitative comparison of segmentation performance and computational complexity evaluated at an input resolution of $1024 \times 1024$.
The best result is highlighted in bold, and the second-best result is underlined.}
\label{tab:quantitative_comparison}
\begin{tabular}{lcccc|cc}
\toprule
Model (Year) & Dice & Recall & Precision & HD95 & Params (M) & FLOPs (G) \\
\midrule
U-Net~\cite{Ronneberger:2015} (2015)        & 0.9467 & 0.9573 & 0.8898 & 4.6064 & 31.04 & 1751.61 \\ 
DeepLabV3+~\cite{deeplabv3+} (2018)   & 0.9585 & 0.9615 & \underline{0.9566} & 0.0156 & 2.67  & 295.23  \\
TransUNet~\cite{transunet} (2021)   & 0.9587  & \underline{0.9711}  & 0.9478  & 2.5238  & 93.74 & 1044.91  \\
UNETR~\cite{hatamizadeh2022unetr} (2022)      & 0.8638  & 0.8876  & 0.8492  & 19.1559  & 92.02 & 2460.28 \\
nnU-Net~\cite{isensee2021nnu} (2024)    & 0.9006  & 0.8704  & 0.9453  & 387.1896  & 58.65  & 367.97   \\
MedSAM~\cite{medsam} (2024)     & \underline{0.9602} & 0.9648 & 0.9564 & \underline{0.0146} & 90.49 & 743.98 \\

\midrule 
Ours        & \textbf{0.9635}  & \textbf{0.9734}  & \textbf{0.9590}  & \textbf{0.0138}  & 90.49  & 743.98      \\
\bottomrule
\end{tabular}
\end{table}

\begin{figure}[ht]
\centering
\includegraphics[width=\textwidth]{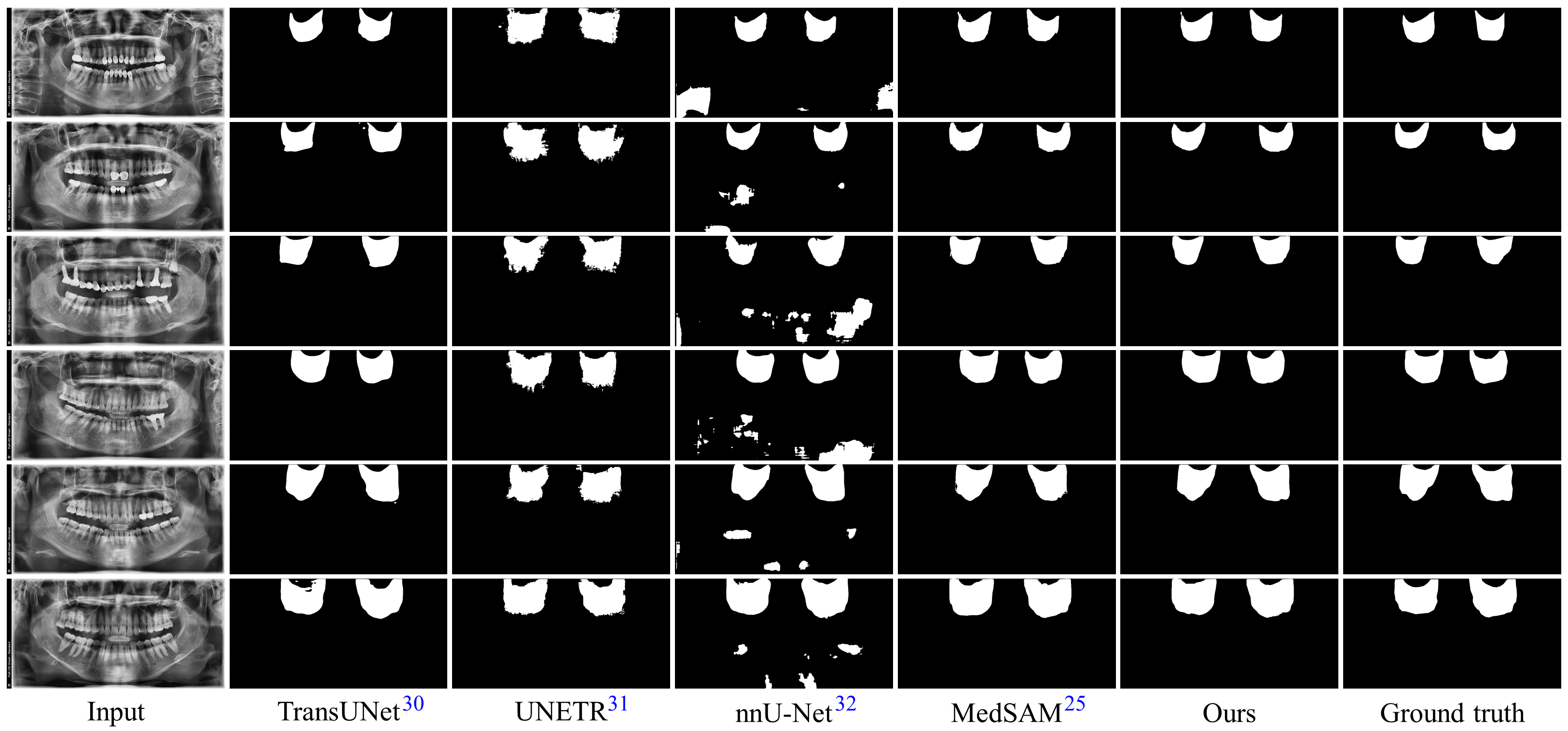}
\caption{Qualitative comparison of maxillary sinus segmentation results obtained using different methods.
From left to right, the columns show the input panoramic X-ray image, TransUNet, UNETR, nnU-Net, MedSAM, the proposed method, and the ground truth mask.
The proposed method produces smoother and more anatomically consistent boundaries compared with the other methods.}
\label{fig:result}
\end{figure}

\subsection*{Qualitative Evaluation}

Figure~\ref{fig:result} presents the  qualitative comparison of the result of maxillary sinus segmentation obtained using different methods.
TransUNet is generally capable of localizing the maxillary sinus region and capturing its bilateral structure; however, its predicted boundaries are often coarse and fail to accurately follow the true anatomical contours, particularly along complex sinus margins.
UNETR produces particularly rough and unstable boundaries, while nnU-Net frequently predicts anatomically irrelevant distant background regions, resulting in severe boundary errors.
These qualitative failure modes are consistent with their lower Dice scores and markedly increased HD95 values reported in Table~\ref{tab:quantitative_comparison}, highlighting their limited robustness in boundary-sensitive segmentation tasks.
MedSAM, as a foundation model pretrained on large-scale segmentation data, generates predictions that more closely resemble the ground truth in terms of overall shape and spatial localization, demonstrating improved anatomical awareness compared with the other baseline methods; nevertheless, minor boundary inaccuracies and local discontinuities remain evident when applied directly to panoramic radiographs.
In contrast, the proposed method leverages knowledge distillation together with refined pseudo label supervision to further correct residual errors and achieve smoother, more anatomically consistent boundaries while effectively suppressing spurious background predictions.
As a result, the proposed framework achieves the closest visual agreement with the ground truth among all compared methods.
Overall, these qualitative observations align well with the quantitative results, particularly the improved Dice score and reduced HD95, demonstrating that the proposed approach effectively enhances both segmentation accuracy and boundary fidelity beyond what is achieved by the foundation model alone.

\subsection*{Ablation study}

We conducted ablation study to demonstrate the effectiveness of the proposed algorithm, and the results are presented in Table~\ref{tab:ablation1}.
The first group of experiments analyzes the impact of incorporating unlabeled data and KD on segmentation performance. When only labeled data were used without KD, the model achieved the Dice score of 96.02\%, recall of 96.48\%, precision of 95.64\%, and HD95 of 0.0146.
Applying the conventional KD proposed by Hinton et al.~\cite{Hinton:2015} without HD95-based weighting using only labeled data led to a slight improvement in Dice to 96.14\% and a substantial increase in recall to 97.45\%, indicating that leveraging the teacher’s predictive distribution alone can enhance region-level detection performance.
However, when unlabeled data were additionally incorporated using the same unweighted KD loss, recall further increased to 97.46\%, whereas the Dice score slightly decreased to 96.08\%.
This degradation suggests that pseudo labels generated for unlabeled data contain a substantial amount of noise and that directly using such noisy pseudo labels can negatively affect segmentation quality.
This observation highlights the necessity of  pseudo label refinement when exploiting unlabeled data.

The second group of experiments evaluates the effectiveness of the proposed refinement network SinusCycle-GAN under a setting where both labeled and unlabeled data are utilized, with a particular focus on the role of the CBAM.
Without CBAM, the SinusCycle-GAN achieved the Dice score of 96.13\% and the highest recall among all configurations, reaching 97.66\%.
This result demonstrates that refinement alone effectively mitigates performance degradation caused by noisy pseudo labels and restores region coverage.
This result directly addresses the Dice drop observed in the first group of experiments when unlabeled data were introduced.
When CBAM was incorporated into SinusCycle-GAN, the Dice score further improved to 96.21\%, precision increased to 95.13\%, and HD95 decreased to 0.0139.
These improvements indicate that CBAM enhances boundary-sensitive feature representation, enabling more accurate and stable boundary refinement beyond coarse structural correction.

The final group of experiments presents the performance of the complete framework, which combines the weighted KD loss based on the HD95 with the SinusCycle-GAN.
Applying weighted KD alone resulted in the Dice score of 96.13\%, recall of 97.08\%, precision of 95.31\%, and HD95 of 0.0141.
When the refinement network was further integrated, the proposed method achieved the best overall performance, with the Dice score of 96.35\%, the highest precision of 95.90\%, and the lowest HD95 of 0.0138.
These results demonstrate that weighting the distillation process improves the reliability of knowledge distillation, while SinusCycle-GAN effectively corrects noisy pseudo labels, allowing the model to fully benefit from unlabeled data.
The ablation study confirms that each component contributes complementarily to the final performance, leading to more accurate and boundary-consistent maxillary sinus segmentation.

\begin{table}[h]
\centering
\caption{Ablation study of the contributions of unlabeled data, weighted knowledge distillation, and pseudo label refinement. Each row incrementally adds a component to the baseline model to evaluate its impact on segmentation accuracy and boundary quality.}
\label{tab:ablation1}
\begin{tabular}{c c c c c c c}
\toprule
Unlabeled & KD & Refinement
& Dice & Recall & Precision & HD95 \\
\midrule

&  &  & 0.9602 & 0.9648 & 0.9564 & 0.0146 \\
& w.o. $w_i$ &  & 0.9614 & 0.9745 & 0.9496 & 0.0142 \\
\checkmark & w.o. $w_i$ &  & 0.9608 & 0.9746 & 0.9484 & 0.0143 \\
\midrule

\checkmark &  & w.o. CBAM & 0.9613 & \textbf{0.9766} & 0.9475 & 0.0141 \\
\checkmark &  & \checkmark & 0.9621 & 0.9740 & 0.9513 & 0.0139 \\
\midrule

\checkmark & \checkmark &  & 0.9613 & 0.9708 & 0.9531 & 0.0141 \\
\checkmark & \checkmark & \checkmark & \textbf{0.9635} & 0.9734 & \textbf{0.9590} & \textbf{0.0138} \\
\bottomrule
\end{tabular}
\end{table}

The proposed framework involves heuristic weighting parameters that control the relative contribution of supervised and unsupervised loss terms.
Specifically, a single hyperparameter $\alpha \in [0,1]$ is used to balance supervised and unsupervised learning by assigning a weight of $\alpha$ to the supervised loss and a complementary weight of $1-\alpha$ to the unsupervised loss.
To examine the effect of these parameters, we conducted an ablation study by varying the weighting ratio between the supervised and unsupervised losses while keeping all other settings fixed.
The quantitative results are summarized in Table~\ref{tab:ablation2}.
As shown in the table, segmentation performance remains relatively stable across different weighting ratios, indicating that the proposed framework is robust to the choice of loss weighting.
The best overall performance is achieved when the supervised and unsupervised losses are equally weighted, yielding the highest Dice score and precision, as well as the lowest HD95 value.
This result suggests that a balanced contribution of labeled and unlabeled data enables effective utilization of pseudo labels while maintaining reliable supervision from ground truth annotations.
In contrast, when the unsupervised loss becomes overly dominant, a slight degradation in segmentation accuracy and boundary quality is observed, which can be attributed to the increased influence of residual noise in pseudo labels.

\begin{table}[h]
\centering
\caption{Effect of varying the weighting ratio between supervised and unsupervised loss terms on segmentation performance. The table reports segmentation results under different ratios of supervised and unsupervised loss terms to analyze the sensitivity of the proposed framework to heuristic weighting parameters.}
\label{tab:ablation2}
\begin{tabular}{c | c c c c}
\toprule
$\alpha$ & Dice & Recall & Precision & HD95  \\
\midrule
0.1 & 0.9618 & 0.9735 & 0.9493 & 0.0143 \\
0.3 & 0.9619 & \textbf{0.9752} & 0.9481 & 0.0141 \\
0.5 & \textbf{0.9635}  & 0.9734  & \textbf{0.9590}  & \textbf{0.0138} \\
0.7 & 0.9624 & 0.9750 & 0.9491 & 0.0140 \\
0.9 & 0.9625 & 0.9744 & 0.9500 & 0.0140 \\

\bottomrule
\end{tabular}
\end{table}

\begin{figure}[ht]
\centering
\includegraphics[width=\linewidth]{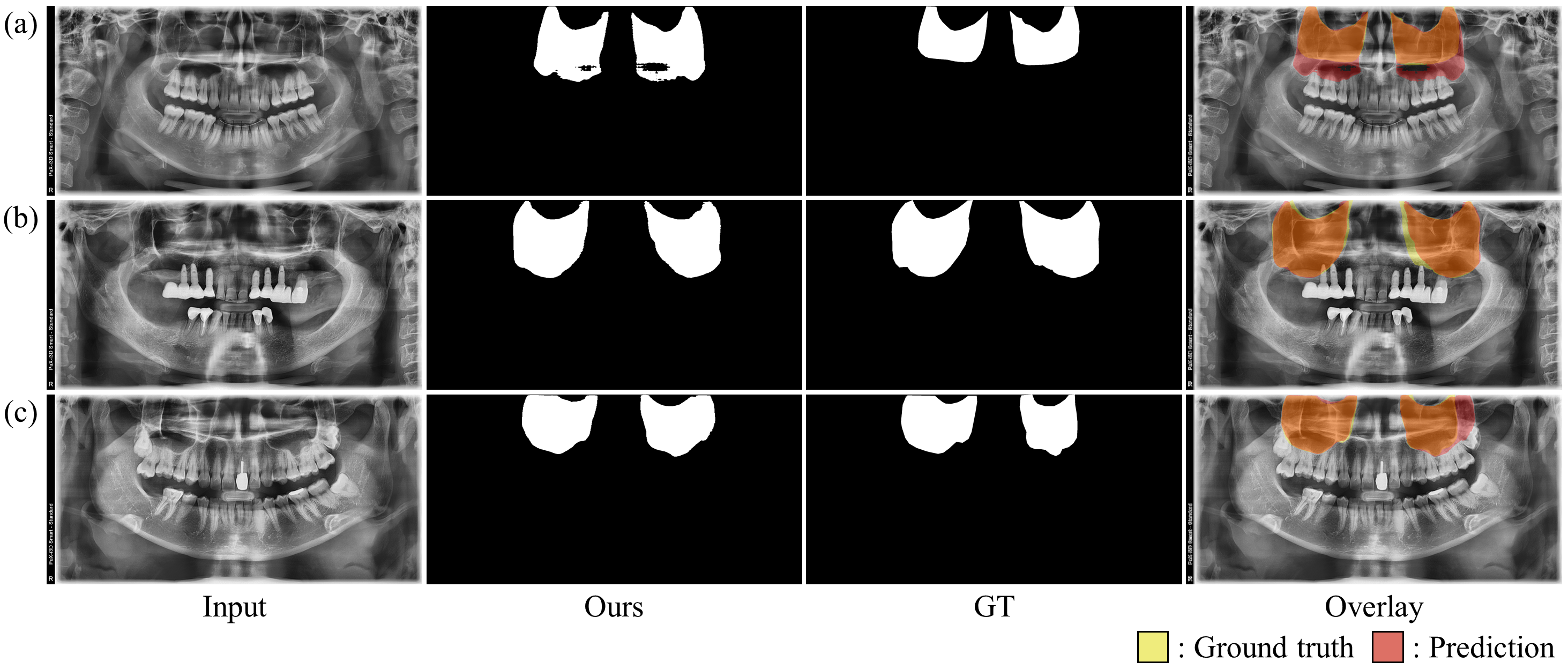}
\caption{Representative failure cases of the proposed method for maxillary sinus segmentation in panoramic X-ray images.
(a) internal holes. (b) underestimation of the ground truth region. (c) overestimation of the ground truth region.}
\label{fig:failure}
\end{figure}

\section*{Failure case}
Figure~\ref{fig:failure} illustrates representative failure cases of the proposed method, revealing limitations primarily associated with the inherent characteristics of panoramic radiography, including reduced image contrast, structural superposition, and anatomical variability.
Across the presented examples, inaccurate boundary modeling manifests in several forms of segmentation errors, including internal discontinuities within the sinus region, incomplete coverage of the true anatomical extent, and erroneous inclusion of surrounding non-sinus structures.
Specifically, Figure~\ref{fig:failure}(a) demonstrates internal false-negative regions within the maxillary sinus, while Figure~\ref{fig:failure}(b) and Figure~\ref{fig:failure}(c) reflect insufficient and excessive coverage of the sinus region near ambiguous margins.
These error patterns indicate that the proposed framework remains sensitive to low-contrast boundaries and overlapping anatomical structures, where local boundary cues are insufficient for precise delineation.
Nevertheless, such failure cases are observed in a relatively small number of samples.
Despite the inherent limitations of panoramic radiographs and the challenges posed by limited availability of pixel-level annotations, the proposed framework consistently achieves stable and meaningful segmentation improvements across the majority of cases.
These observations suggest that while boundary ambiguity remains a challenge in certain scenarios, the proposed semi-supervised approach effectively mitigates annotation scarcity and provides robust performance in clinically realistic settings.
Incorporating stronger boundary-aware supervision or explicit anatomical priors may further improve robustness in these challenging cases and will be explored in future work.

\section*{Conclusion}
In this paper, we proposed a semi-supervised framework for maxillary sinus segmentation in panoramic X-ray images that jointly leverages a limited amount of labeled data and a large volume of unlabeled data.
To mitigate discrepancies between the teacher and student predictions during knowledge distillation, we introduced a weighted loss function to emphasize boundary-sensitive errors.
In addition, we proposed SinusCycle-GAN, a pseudo label refinement network built upon an unpaired image-to-image translation framework, which incorporates a correction network with an encoder-decoder architecture that includes CBAM to improve the structural accuracy of pseudo labels generated by the teacher.
Experimental results on a clinically acquired dataset demonstrated consistent performance improvements over MedSAM, with improvements of 0.33\%, 0.86\%, and 0.26\% in Dice score, recall, and precision, respectively.
The experimental results suggest that the proposed approach provides a practical and effective solution for dental image analysis scenarios where pixel-level annotation is costly and labor-intensive, highlighting its potential to reduce annotation burden while maintaining high segmentation accuracy.
Future work will focus on enhancing robustness to challenging imaging conditions by incorporating additional structural priors and more advanced boundary-aware learning strategies.

\section*{Data availability}
Data used for this study is available on request to the corresponding author.

\bibliography{sample}

\section*{Author contributions}
J.P. developed the algorithm, conducted the experiments, and wrote the initial manuscript.
J.C. revised the manuscript.
J.P.Y. proposed the conception of the study.
Y.C.P. performed data analysis.
H.G.Y. contributed to data acquisition.
B.D.L. contributed to data acquisition.
S.J.L. revised the manuscript. All authors reviewed the manuscript.

\section*{Funding}
This work was supported by the Institute of Information \& Communications Technology Planning \& Evaluation(IITP)-Innovative Human Resource Development for Local Intellectualization program grant funded by the Korea government(MSIT) (IITP-2025-RS-2024-00439292).
This study was supported by the Korea Institute of Industrial Technology(EH260004).

\section*{Competing interests}
The authors declare no competing interests.

\end{document}